\documentclass[letterpaper, 10 pt, conference]{ieeeconf}  

\IEEEoverridecommandlockouts                              
\usepackage{amsmath,array,graphicx}
\usepackage{graphics} 
\graphicspath{{images/},{images/random_images}}                          
\usepackage[]{subfig}
\usepackage{color}
\usepackage{multirow}
\usepackage{hyperref}
\usepackage{nicefrac}
\begin{document}
\title{\LARGE \bf
Agreeing To Cross: How Drivers and Pedestrians Communicate*}

\author{Amir Rasouli, Iuliia Kotseruba and John K. Tsotsos$^{1}$
\thanks{*This work was supported by the Natural Sciences and Engineering Research Council of Canada (NSERC), the NSERC Strategic Network for Field Robotics (NCFRN), and the Canada Research Chairs Program through grants to JKT.}
\thanks{$^{1}$The authors are with The Department of Electrical Engineering and Computer Science and Center for Vision Research York University, Toronto, Canada.
 {\tt\small $\{aras;yulia\_k;tsotsos\}$@eecs.yorku.ca}}}%

\maketitle
\thispagestyle{empty}
\pagestyle{empty}

\begin{abstract}
The contribution of this paper is twofold. The first is a novel dataset for studying behaviors of traffic participants while crossing. Our dataset contains more than 650 samples of pedestrian behaviors in various street configurations and weather conditions. These examples were selected from approx. 240 hours of driving in the city, suburban and urban roads.
  
The second contribution is an analysis of our data from the point of view of joint attention. We identify what types of non-verbal communication cues road users use at the point of crossing, their responses, and under what circumstances the crossing event takes place.

It was found that in more than $90$\% of the cases pedestrians gaze at the approaching cars prior to crossing in non-signalized crosswalks. The crossing action, however, depends on additional factors such as time to collision, explicit driver's reaction or structure of the crosswalk. 
\end{abstract}

\section{INTRODUCTION}

The fascination with autonomously driving vehicles goes as far back as mass production of early automobiles. Since the early 1920s the automotive industry has witnessed numerous attempts to achieve full autonomy in the form of radio signal controlled cars \cite{Kroger2016}, wire following vehicles \cite{Mann1958}, lane detection and car following \cite{Dickmanns1990} and, in more recent works, the cars that can fully autonomously drive roads under certain conditions \cite{SebastianThrun2006}.

Despite such success stories in autonomous control systems, designing fully autonomous vehicles suitable for urban environments still remains an unsolved problem. Aside from challenges associated with developing suitable infrastructure \cite{Friedrich2016} and regulating the autonomous behaviors \cite{Gasser2016}, one of the major dilemmas faced by autonomous vehicles is to how to communicate with other road users in a chaotic traffic scene \cite{Knight2015a}. In addition to official rules that govern the flow of traffic, humans often rely on some form of informal rules resulting from non-verbal communication among them and anticipation of the other traffic participants' intentions. For instance, pedestrians intending to cross a street where there is no stop sign or traffic signal often establish eye contact with the driver to ensure that the approaching car will stop for them. Other forms of non-verbal communication such as hand gestures or body posture are also used to resolve ambiguities in typical traffic situations. Furthermore, the characteristics of a road user (e.g. age and gender), the physical environment (the structure of the crosswalk, weather, etc.) and even cultural differences make estimating the intention of traffic participants particularly challenging \cite{Wolf2016}.

Our contribution in the proposed work is twofold. First, we introduce a novel visual dataset for detection and analysis of pedestrians' behaviors while crossing (or attempting to cross) the street under various conditions. We call this dataset Joint Attention in Autonomous Driving (JAAD). Then we present some of our findings regarding the course of actions taken and non-verbal cues used by pedestrians in different crossing scenarios. We show that the crossing behavior can be influenced by various contextual elements such as crossway structure, driver's behavior, distance to the approaching vehicles, etc.
  
\begin{figure}[!tbtp]
\centering
\includegraphics[width=1\columnwidth]{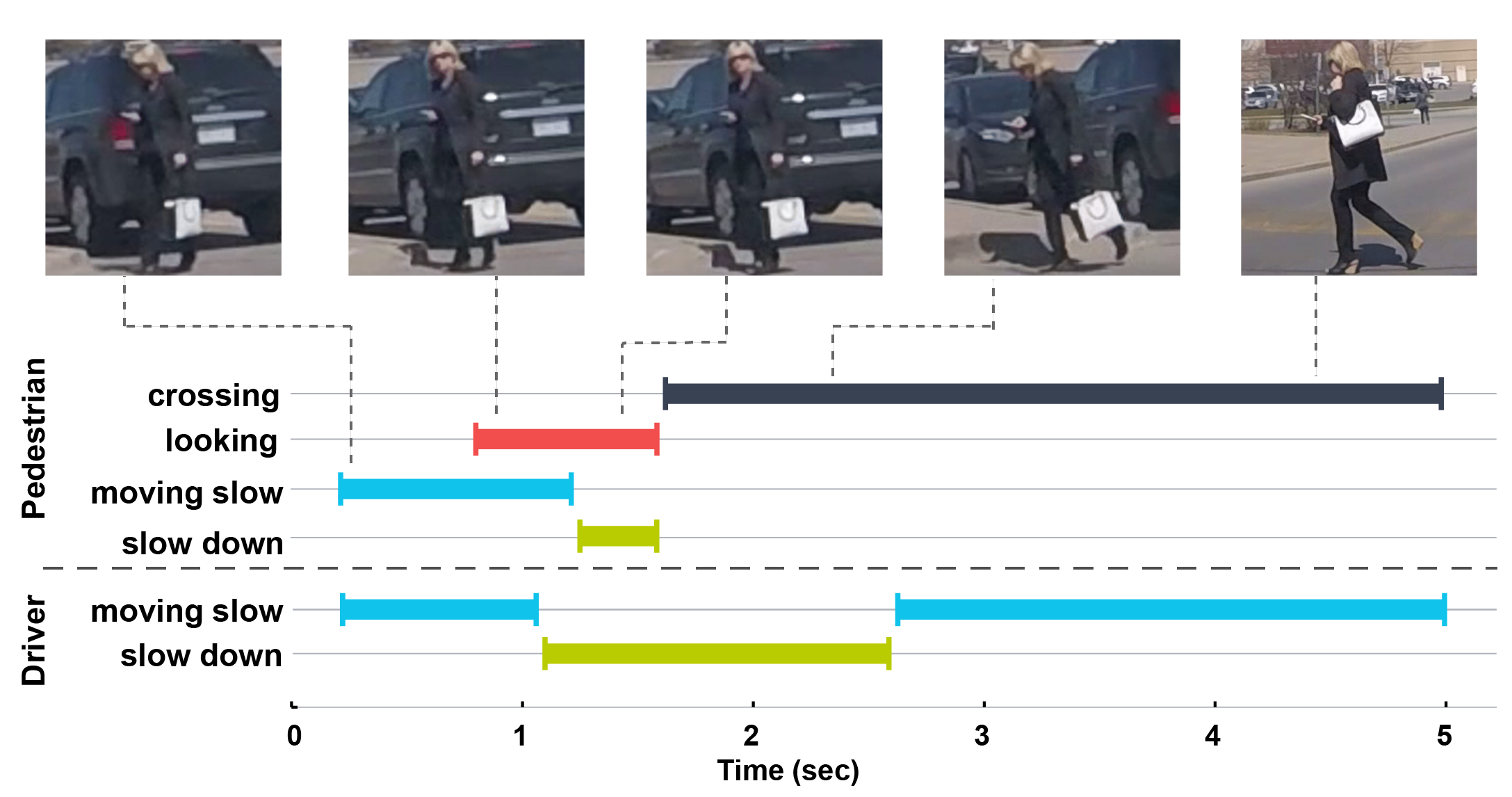} 
\caption{An overview of joint attention in crossing. The timeline of events is recovered from the behavioral data and shows a single pedestrian crossing the parking lot. Initially, the driver is moving slow and, as he notices the pedestrian ahead, slows down to let her pass. At the same time the pedestrian crosses without looking first then turns to check if the road is safe, and, as she sees the driver yielding, continues to cross.}
\label{fig:behavior}
\end{figure}
   
\section{Related Works}

\subsection{Studies of driver and pedestrian interaction}
Numerous psychological studies examined the behaviors of drivers and pedestrians before crossing events. Usually, the following aspects are considered: the likelihood of the driver yielding (\cite{Sun2002, Salamati2013, Gueguen2015}), driver awareness of the pedestrian \cite{Lee2009, Fukagawa2013} and pedestrian's decision making \cite{Tom2011, Sun2015}. Multiple factors affecting these behaviors have been identified: vehicle speed and time to collision (TTC) (\cite{Lubbe2015, Schmidt2009}), size of the gap between the vehicles \cite{Wang2010}, geometry and other features of the road (signs and delineation) \cite{Tom2011}, weather conditions \cite{Sun2015}, crossing conditions (whether pedestrian is crossing from a standstill or walking), number of pedestrians crossing \cite{Wang2010}, gender and age of the drivers and pedestrians \cite{Tom2011}, eye contact between the pedestrian and the driver (\cite{Gueguen2015, Ren2016}), etc. 

Typically, the interactions between the traffic participants are treated mechanistically. For instance, TTC takes into account the speed of the vehicle and distance to the pedestrian and is thought to affect his/her crossing behavior (\cite{Oudejans1996, Schmidt2009, Du2013, Lubbe2015}). 

However, several recent studies show that non-verbal communication is also important for determining the intentions of traffic participants. For example, drivers are more likely to yield if they are looked at by the pedestrian waiting to cross (\cite{Gueguen2015, Ren2016}). 

In a psychological experiment by Schmidt \textit{et al.} \cite{Schmidt2009} participants were unable to correctly evaluate pedestrians' crossing intentions based only on the trajectories of their motion, suggesting that parameters of body language (posture, leg and head movements) are valuable cues.

In computer vision and robotics, passive approaches are prevalent for predicting pedestrians' actions during the crossing. These works mainly look at the dynamic factors in the scene such as pedestrians' trajectories \cite{Bandyopadhyay2013} and velocities \cite{Pellegrini2009} or try to predict the changes in the behavior of pedestrians crossing as a group \cite{Choi2014}.

In more recent works, the pedestrian's body language is used as a means of predicting behavior \cite{Kooij2014, Schulz2015}. In these works, head orientation is associated with the pedestrian's level of awareness, however, the learning is crude and the context is not taken into account. For instance, driver's reaction or vehicle's speed as well as the structure of the crossway such as presence of a traffic signal or width of the street is not considered.

\subsection{Existing Datasets}
There are many datasets for pedestrian detection introduced by the computer vision and robotics communities. To name a few, KITTI \cite{Geiger2013}, Caltech pedestrian  detection benchmark \cite{Dollar2012} and Daimler Pedestrian Benchmark Dataset \cite{Enzweiler2009}. These datasets are accompanied by ground truth information in the form of bounding boxes, stereo information, sensor readings and occlusion tags. 

To the best of our knowledge, there are no datasets facilitating the study of pedestrians' crossing behavior. Most of the data for the relevant psychological studies is collected at select locations and involves direct observation by the researchers on site. Another potential source is data collected for Naturalistic Driving Studies (NDS). These are introduced to eliminate observer's effect and aggregated large volumes of data on everyday driving patterns over an extended period of time. A number of such studies have been launched in the USA \cite{SHRP2,VTTI}, Europe \cite{Barnard2016}, Asia \cite{Uchida2010} and Australia \cite{Williamson2015}. Although these studies produced petabytes of video recordings of everyday driving situations, at present the processing of this data has been focused on identifying crash and near-crash events and factors that caused them. Since access to the raw NDS data is restricted and only general anonymized statistics are available, we conducted a small-scale naturalistic driving study and extracted data on the non-verbal communication occurring between the traffic participants in various situations. The following sections discuss data collection procedure, general statistics and the preliminary results of our study.

\begin{figure*}
\centering
\subfloat[crossing events]{\includegraphics[width=.85\textwidth]{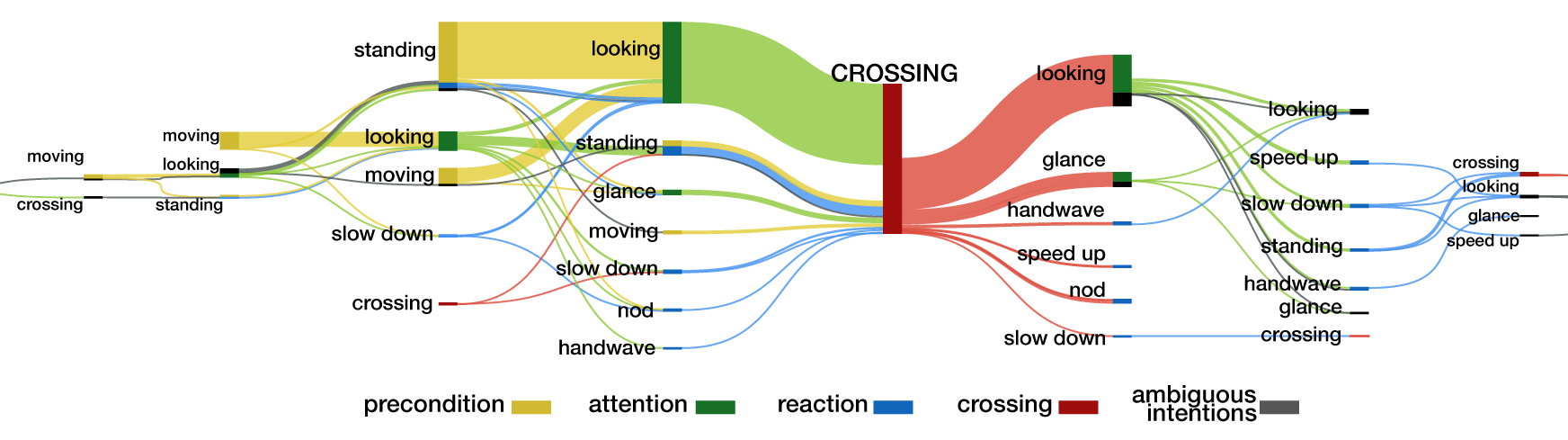}
\label{fig:crossing_chart}}\\
\subfloat[no crossing events]{\includegraphics[width=.85\textwidth]{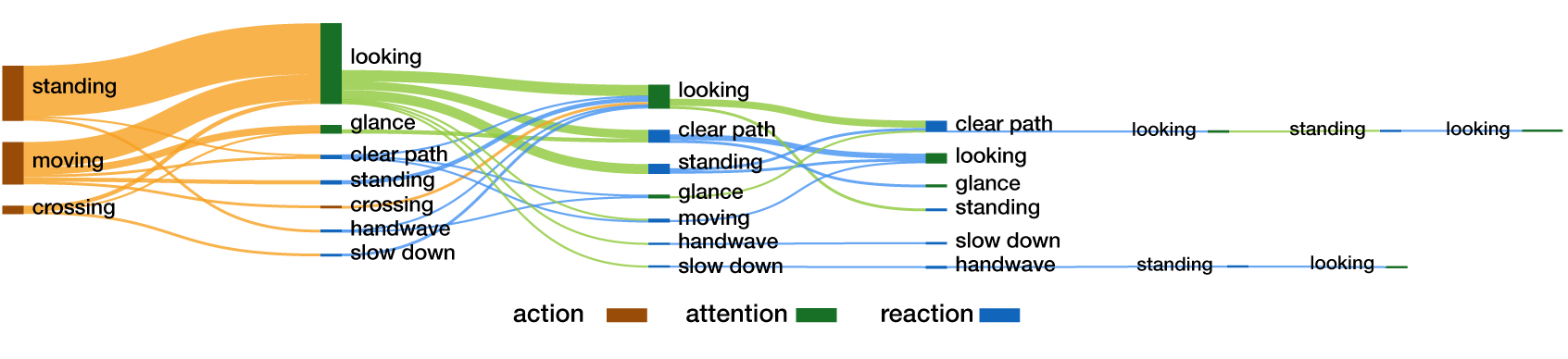}
\label{fig:no_crossing_chart}} 
\caption{ Joint attention motifs of pedestrians. Diagram a) shows a summary of 345 sequences of pedestrians' actions before and after crossing. Diagram b) shows 92 sequences of actions when pedestrians did not cross. Vertical bars represent actions color-coded as the \textit{precondition} to crossing, \textit{attention}, \textit{reaction} to driver's actions, \textit{crossing} or \textit{ambiguous} actions. Curved lines between the bars show connections between consecutive actions. The thickness of lines reflects the frequency of the action in the 'crossing' or 'non-crossing' subset. The sequences longer than 10 actions (e.g. when the pedestrian hesitates to cross) are extremely rare and are not shown.}
\end{figure*}

\section{The JAAD Dataset}
The JAAD dataset\footnote{\href{http://data.nvision2.eecs.yorku.ca/JAAD\_dataset/}{http://data.nvision2.eecs.yorku.ca/JAAD\_dataset/}. Ethics certificate \# 2016-203 from York University. } was created to study the behavior of traffic participants. The data consists of 346 high-resolution video clips (5-15s) showing various situations typical for urban driving. These clips were extracted from approx. 240 hours of driving videos collected in several locations. Two vehicles equipped with wide-angle video cameras were used for data collection (Table \ref{table_sample_locations}). Cameras were mounted inside the cars in the center of the windshield below the rear view mirror.

\begin{table}[!hbtp]
\caption{Properties of the samples in the database.}
\vspace*{-\baselineskip}
\label{table_sample_locations}
\begin{center}
\resizebox{\columnwidth}{!}{
\begin{tabular}{c|c|c|c}
{\textbf{\# Clips}} & \textbf{Location} & \textbf{Resolution} & \textbf{Camera Model} \\
\hline
55 & Toronto, Canada & $1920 \times 1080$ & GoPro HERO+\\
\hline
276 & Kremenchuk, Ukraine & $1920 \times 1080 $ & Garmin GDR-35\\
\hline
6 & Hamburg, Germany & $1280 \times 720$ & Highscreen Black Box Connect\\
\hline
5 & New York, USA & $1920 \times 1080 $ & GoPro HERO +\\
\hline
4 & Lviv, Ukraine & $1280 \times 720$ & Highscreen Black Box Connect\\
\hline
\end{tabular}}
\end{center}
\end{table}
\vspace*{-\baselineskip} 
The video clips represent a wide variety of scenarios involving pedestrians and other drivers. Most of the data is collected in urban areas (downtown and suburban), only a few clips are filmed in rural locations. The samples cover a variety of situations such as pedestrians crossing individually, or as a group, pedestrians occluded by objects, walking along the road and many more. The dataset contains fewer clips of interactions with other drivers, most of them occur in uncontrolled intersections, in parking lots or when another driver is moving across several lanes to make a turn.

The videos are recorded during different times of the day, and under various weather and lighting conditions. Some of them are particularly challenging, for example, sun glare. The weather also can impact the behavior of road users, for example, during the heavy snow or rain people wearing hooded jackets or carrying umbrellas may have limited visibility of the road. Since their faces are obstructed it is also harder to tell if they are paying attention to the traffic from the driver's perspective.

We attempted to capture all of these conditions for further analysis by providing two kinds of annotations for the data: bounding boxes and textual annotations. Bounding boxes are provided only for cars and pedestrians that interact with or require the attention of the driver (e.g. another car yielding to the driver, pedestrian waiting to cross the street, etc.). Bounding boxes for each video are written into an XML file with frame number, coordinates, width, height, and occlusion flag.
The textual annotations are created using the BORIS2 software for video observations \cite{Friard2016}. It allows to assign predefined behavior labels to different subjects seen in the video, and can also save some additional data, such as video file id, the location where the observation has been made, etc. (see Fig. \ref{fig:behavior} for an example). 

We save the following data for each video clip: weather, time of the day, age and gender of the pedestrians, location and whether it is a designated crosswalk.

Each pedestrian is assigned a label (pedestrian1, pedestrian2, etc.). We also distinguish between the driver inside the car and other drivers, which are labeled as Driver and car1, car2, etc. respectively. This is necessary for the situations where two or more drivers are interacting. Finally, a range of behaviors is defined for drivers and pedestrians: walking, standing, looking, moving, etc. A more detailed example of textual annotation can be found in \cite{Kotseruba2016}.

\section{The Data}
In our data, we observed high variability in the behaviors of pedestrians at the point of crossing/no-crossing with more than 100 distinct patterns of actions. For instance, Fig. \ref{fig:crossing_chart} shows sequences of actions during the completed crossing scenarios found in the dataset. Two typical patterns, "standing, looking, crossing" and "crossing, looking", cover only half of the situations observed in the dataset. Similarly, in $\nicefrac{1}{3^{rd}}$ of non-crossing scenarios (Fig. \ref{fig:no_crossing_chart}) pedestrians are waiting at the curb and looking at the traffic. Otherwise, the behaviors vary significantly both in the number of actions before and after crossing and in the meaning of particular actions (e.g. standing may be both a precondition and a reaction to driver's actions).

For further analysis we split these behavioral patterns into 9 groups depending on the initial state of the pedestrian and whether the attention or the act of crossing is happening. We list these actions and the number of samples in Table \ref{table_beh_cat}.  Here attention refers to the first moment the pedestrian is assessing the environment and expressing his/her intention to the approaching vehicles, therefore it is considered as a form of non-verbal communication.

Visual attention takes two forms: looking and glancing. Looking refers to the scenarios in which the pedestrian inspects the approaching car (typically for 1 second or longer), assesses the environment and in some cases establishes eye contact with the driver. The other form of attention, glance, usually lasts less than a second and is used to quickly assess the location or speed of the approaching vehicles. Pedestrians glance when they have a certain level of confidence in predicting the driver's behavior, e.g. the vehicle is stopped or moving very slowly or otherwise is sufficiently far away and does not pose any immediate danger.
     
\begin{table*}[!hbtp]
\caption{The behavioral patterns observed in the data.}
\vspace*{-\baselineskip}
\label{table_beh_cat}
\begin{center}
\resizebox{\textwidth}{!}{
\begin{tabular}{c|c|c}
{\textbf{Behavior Sequence}} & \textbf{Meaning} & \textbf{Number of Samples} \\
\hline
Crossing & The pedestrian is observed at the point of crossing and no attention is taking place & $152$\\
\hline
Crossing + Attention & The pedestrian is observed at the point of crossing and some form of attention is occurred & $64$\\
\hline
Crossing + Attention + Reaction & The pedestrian is observed at the point of crossing and some form of attention is occurred and the pedestrian changes behavior & $29$\\
\hline
PreCondition + Crossing & The pedestrian is walking/standing and crosses without paying attention & $37$ \\
\hline
Precondition + Attention + Crossing & The pedestrian is walking/standing and crosses after paying attention & $160$ \\
\hline
Precondition + Attention + Reaction + Crossing & The pedestrian is walking/standing, pays attention and changes behavior prior to crossing & $64$ \\
\hline
Action & The pedestrian is walking/standing and his/her intention is ambiguous & $56$ \\
\hline
Action + Attention & The pedestrian is about to cross and pays attention & $43$ \\
\hline
Action + Attention + Reaction & The pedestrian is about to cross, pays attention and responds & $49$ \\
\hline
Total & & $654$ \\
\hline
\end{tabular}}
\end{center}
\end{table*}

 \section{Observations and Analysis}
Our data contains various scenarios in which pedestrians are observed during or prior to crossing. Two categories from Table \ref{table_beh_cat}, \textit{crossing} and \textit{action}, are omitted from the analysis. Since these crossing scenarios do not demonstrate the full crossing event, it is difficult to assess the behavior of the pedestrians at the point of crossing. As for the \textit{action} cases the intentions of the pedestrians are ambiguous. For example, pedestrians are not approaching the curb or are standing far away from the crossway.
\subsection{Forms of non-verbal communication}
In the course of a crossing event, pedestrians often use different forms of non-verbal communication (in more than $90$\% of the cases in our dataset). The most prominent signal to transmit the crossing intention is looking ($90$\%) or glancing ($10$\%) towards the coming traffic. Other forms of communication are rarer, e.g. nodding (as a form of gratitude and acknowledgement) and hand gesture (as a form of gratitude or yielding), and are usually performed in response to the driver's action.

The pedestrians' response to the communication is not always explicit and is often realized as a change in their behavior. For instance, when a pedestrian slows down or stops it could be an indicator of noticing the vehicle approaching or driver not yielding. Table \ref{table_comms_responses} summarizes the forms of communication and responses observed in the data. In this table we distinguish between the primary and secondary occurrence of attention. The primary attention is the first instance when the pedestrian inspects the environment prior to crossing. The secondary attention refers subsequent inspection of the environment or checking the traffic while crossing.

\begin{table}[!hbtp]
\caption{Forms of pedestrians communication and response. Primary (PO) and secondary occurrence (SO) of attention.}
\vspace*{-\baselineskip}
\label{table_comms_responses}
\begin{center}
\begin{tabular}{cc|c|c}
& & Form of Communication & Number  of Occurrences\\
\hline
\multicolumn{1}{ c}{\multirow{4}{*}{attention}} &
\multicolumn{1}{|c|}{\multirow{2}{*}{PO}} 	   & looking & 328\\  \cline{3-4}
 & \multicolumn{1}{|c|}{} 					   & glance & 37\\   \cline{2-4}
 & \multicolumn{1}{|c|}{\multirow{2}{*}{SO}}	   & looking & 106\\  \cline{3-4}
 & \multicolumn{1}{|c|}{} 					   & glance  & 19\\ \cline{1-4}
\multicolumn{2}{ c|}{\multirow{6}{*}{response}} & stop & 71\\ \cline{3-4}
\multicolumn{2}{ c|}{}						   & clear path & 29\\ \cline{3-4}
\multicolumn{2}{ c|}{}						   & slow down & 24\\ \cline{3-4}
\multicolumn{2}{ c|}{}						   & speed up & 14\\ \cline{3-4}
\multicolumn{2}{ c|}{}						   & hand gesture & 13\\ \cline{3-4}
\multicolumn{2}{ c|}{}						   & nod & 11\\ \hline	 

\end{tabular}
\end{center}
\end{table}
\vspace*{-\baselineskip}

\subsection{Attention occurrence prior to crossing}
As mentioned earlier there are scenarios in which pedestrians do not pay attention to the moving traffic. To investigate the probability of attention occurrence, one important factor to consider is TTC or how long it takes the approaching vehicle to arrive at the position of the pedestrian, given that they maintain their current speed and trajectory.

\begin{figure}[!tbtp]
\centering
\includegraphics[width=0.23\textwidth]{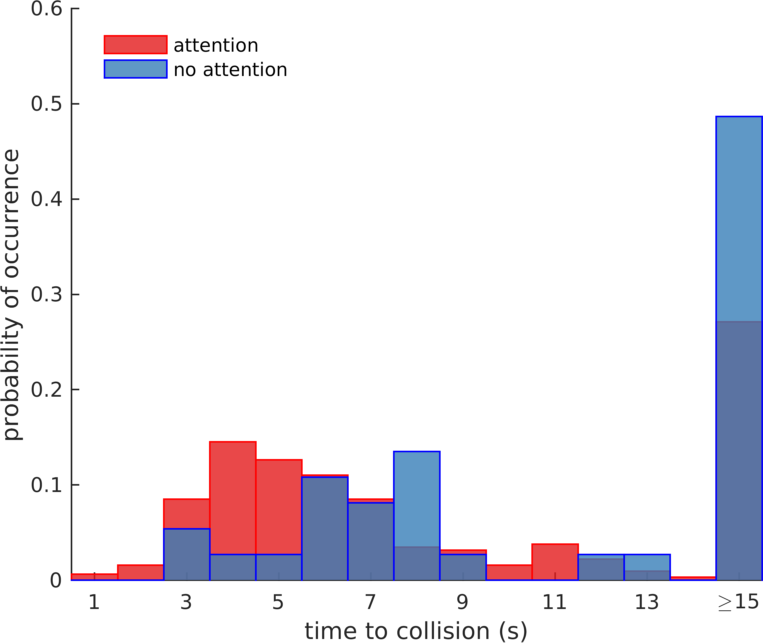}
\caption{Relationship between TTC and probability of attention occurring prior to crossing.}
\label{fig:TTC_attention}
\end{figure}

The relationship between attention occurrence and TTC is illustrated in Fig. \ref{fig:TTC_attention}. Crossing without attention comprises only about $10$\% of all crossing scenarios out of which more than $50$\% of the cases occurred when TTC is above 10s (including situations where the approaching vehicle is stopping). There is also no cases of crossing without attention when TTC is less than $2s$. 

The context in which the crossing takes place also plays a role in crossing behavior. The context can be described by factors such as the weather conditions, street structure and driver's reaction. Since analyzing all these factors is beyond the scope of this paper, here we only look at the effect of the street structure.

\begin{figure}[!tbtp]
\centering
\subfloat[]{\includegraphics[width=0.23\textwidth]{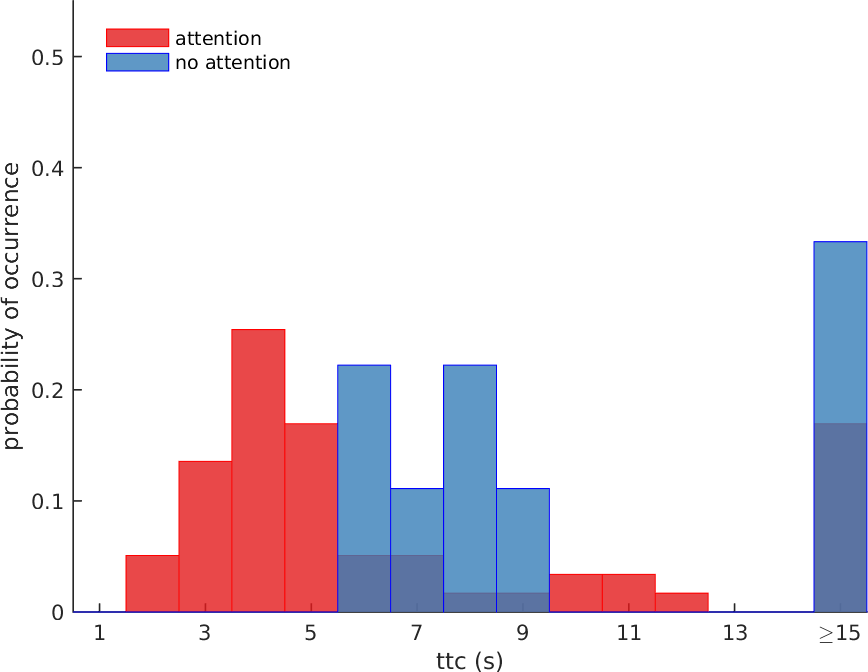} 
\label{fig:ttc_lanes:below3}}
\subfloat[]{\includegraphics[width=0.23\textwidth]{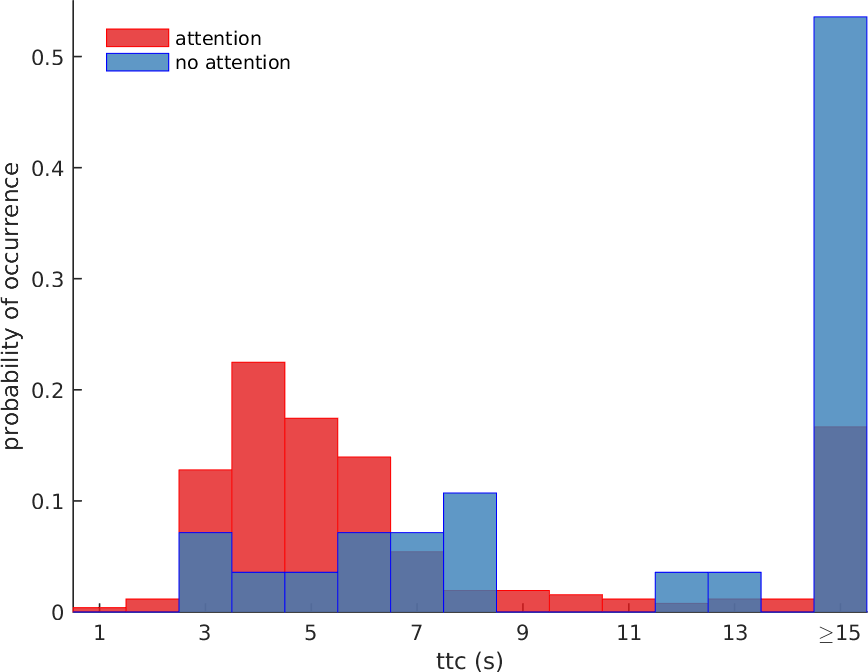}
\label{fig:ttc_lanes:above3}}\\
\caption{The pedestrian attention frequency at a) designated and  b) non-designated crosswalks.}
\label{fig:ttc_des_att}
\end{figure}

There are two factors that characterize a crosswalk: whether it is designated (there is a zebra crossing or traffic signal) and its width (measured as the number of lanes). In our samples, crossing without attention only happened in non-designated crosswalks when TTC was higher than $6$ seconds (see Fig. \ref{fig:ttc_des_att}).

The full crossing events happen in street with widths ranging from 1 (narrow one-way streets) to 4 lanes (main streets).

\begin{figure}[!tbtp]
\centering
\subfloat[]{\includegraphics[width=0.23\textwidth]{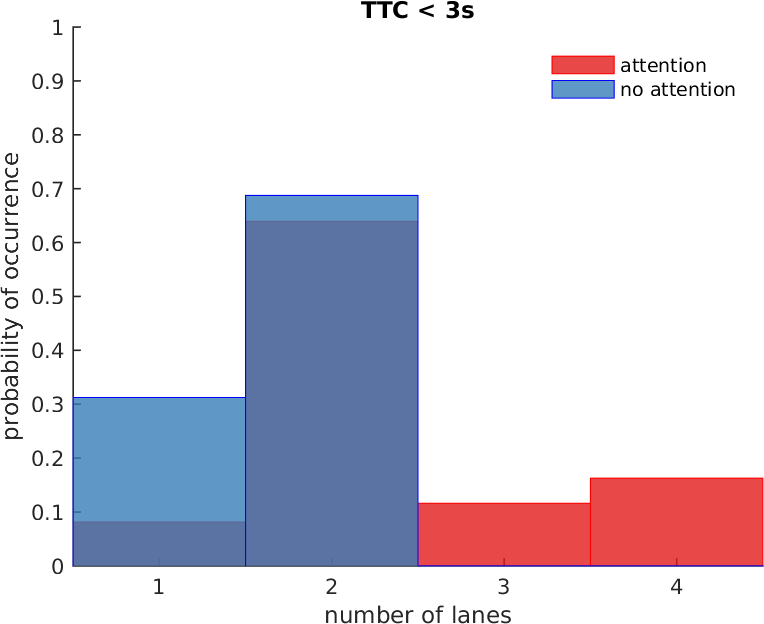} 
\label{fig:ttc_lanes:below3}}
\subfloat[]{\includegraphics[width=0.216\textwidth]{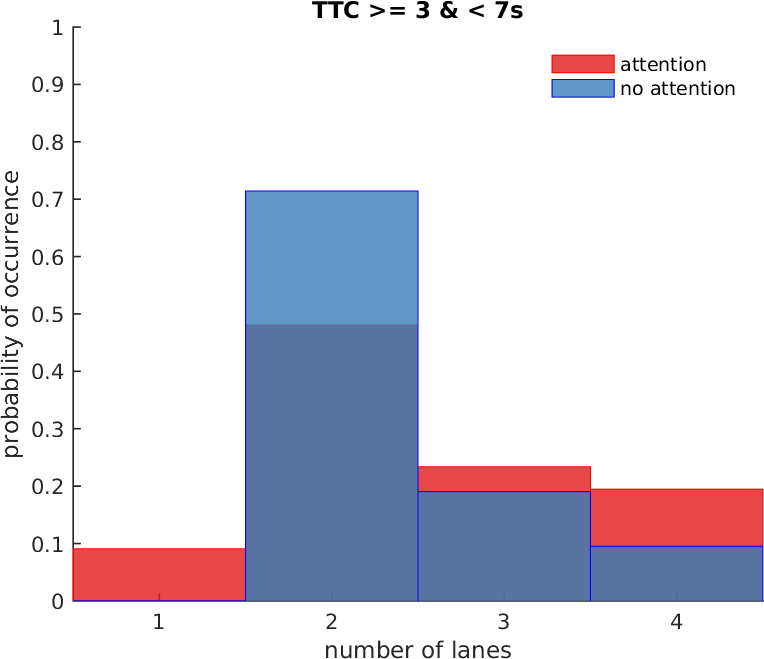}
\label{fig:ttc_lanes:above3}}\\
\subfloat[]{\includegraphics[width=0.23\textwidth]{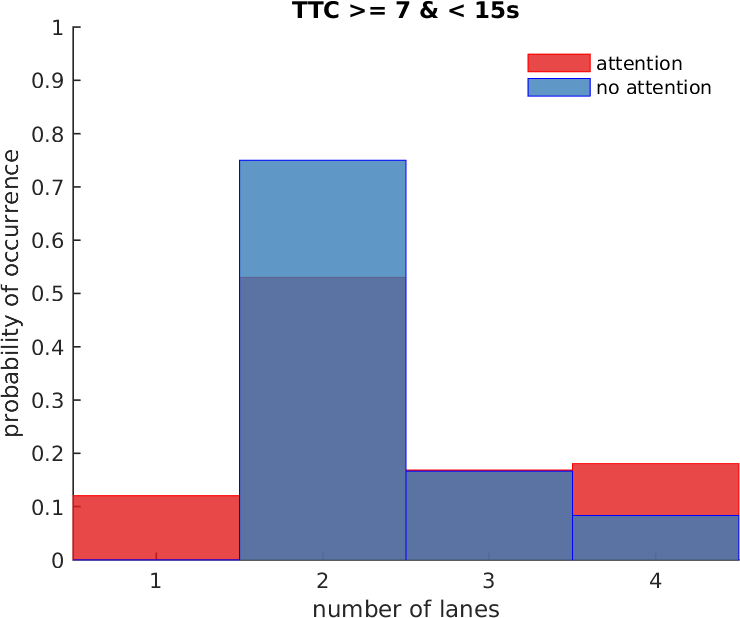}
\label{fig:ttc_lanes:above7}}
\subfloat[]{\includegraphics[width=0.23\textwidth]{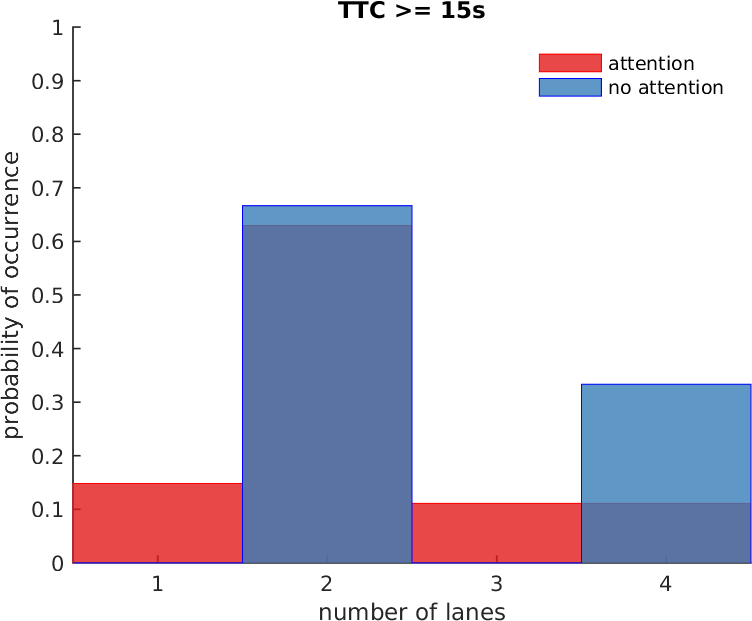}
\label{fig:ttc_lanes:above15}}\\
\caption{Attention occurrence with respect to the number of lanes.}

\label{fig:ttc_lanes}
\end{figure}

We report on the data by dividing the results into 4 intervals with respect to the TTC values and in each category, we group them based on the number of lanes (see Fig. \ref{fig:ttc_lanes}). As illustrated, when TTC is below 3s there is no occurrence of crossing without attention in streets wider than 2 lanes. In fact, only $18$\% of the crossings happened in streets wider than 2 lanes.

\begin{figure}[!tbtp]
\centering
\includegraphics[width=0.25\textwidth]{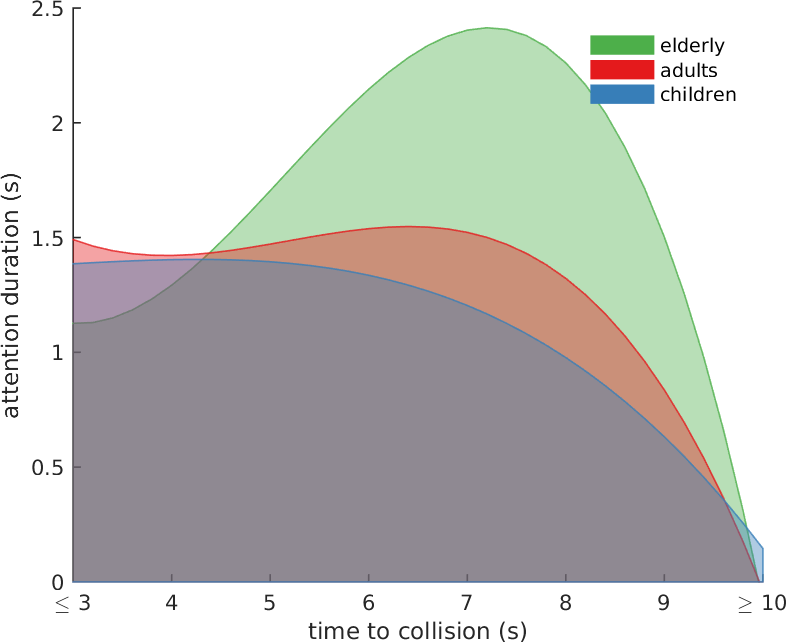}
\caption{Average duration of the pedestrian's attention prior to crossing based on TTC for different age groups.}
\label{fig:TTC_att_duration}
\end{figure}

The duration of attention or how fast pedestrians tend to begin crossing from the moment they gaze at the approaching car also may vary. As illustrated in Fig. \ref{fig:TTC_att_duration}, the duration of looking depends on time to collision. The further away the vehicle is from the pedestrians, the longer it will take them to assess the intention of the driver, hence they will attend longer. The gaze duration increases up to a maximum safe TTC threshold (from 7s for adults up to 8s for elderly) after which it dramatically declines when the vehicle is either far away or stopped. In addition, the elderly pedestrians in comparison to adults and children tend to be more conservative and spend on average about 1s longer on looking prior to crossing. 
   
\subsection{Crossing action post attention occurrence}
Although the pedestrian's head orientation and attentive behavior are strong indicators of crossing intention, they are not always followed by a crossing event. In addition to TTC, which reflects both the approaching driver's speed and their distance to the contact point, the structure of the street and the driver's reaction can impact the pedestrians level of confidence to cross. 

To investigate this we divide the crosswalks into three categories: \textit{non-designated}, without zebra or traffic signal, \textit{zebra-crossing}, with either zebra or/and a pedestrian crossing sign and \textit{traffic signal} with a signal such as traffic light or stop sign which forces the driver to stop.

\begin{figure}[!tbtp]
\centering
\subfloat[crossing and crosswalk property]{\includegraphics[width=0.23\textwidth]{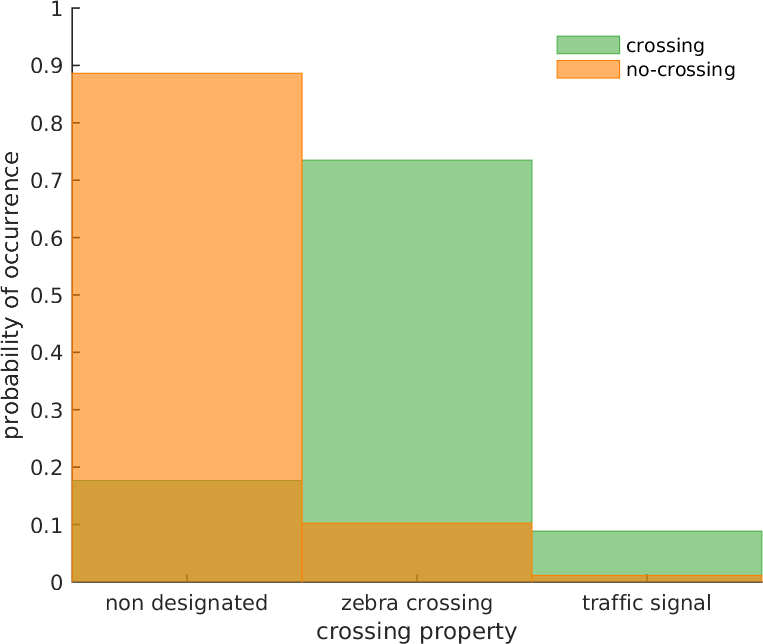} 
\label{fig:crossing:crossing_des}}
\subfloat[non-designated]{\includegraphics[width=0.23\textwidth]{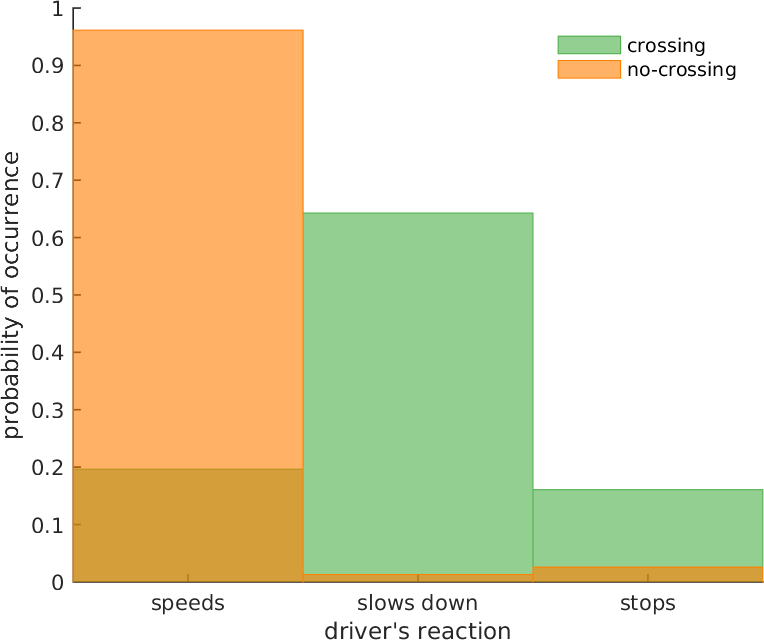}
\label{fig:crossing:driver_reaction_no_des}}\\
\subfloat[zebra crossing]{\includegraphics[width=0.23\textwidth]{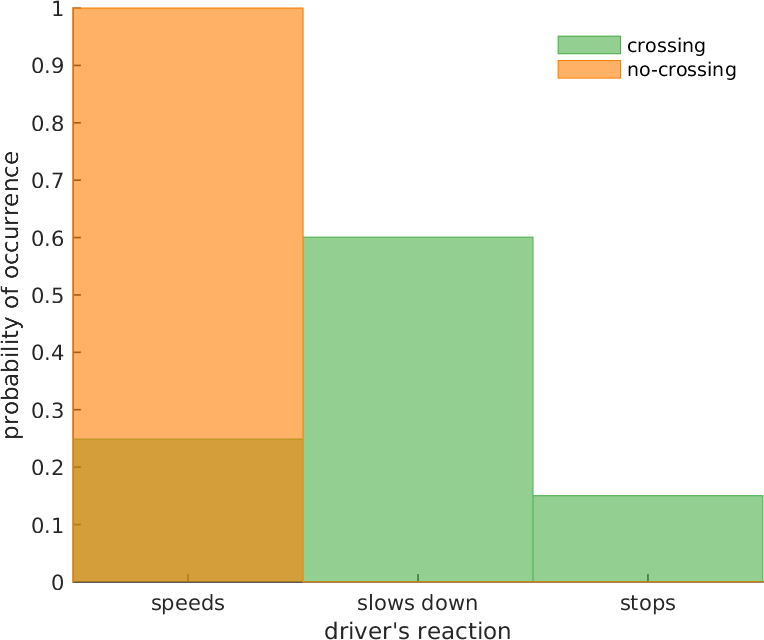}
\label{fig:crossing:driver_reaction_des}}
\subfloat[traffic signal]{\includegraphics[width=0.23\textwidth]{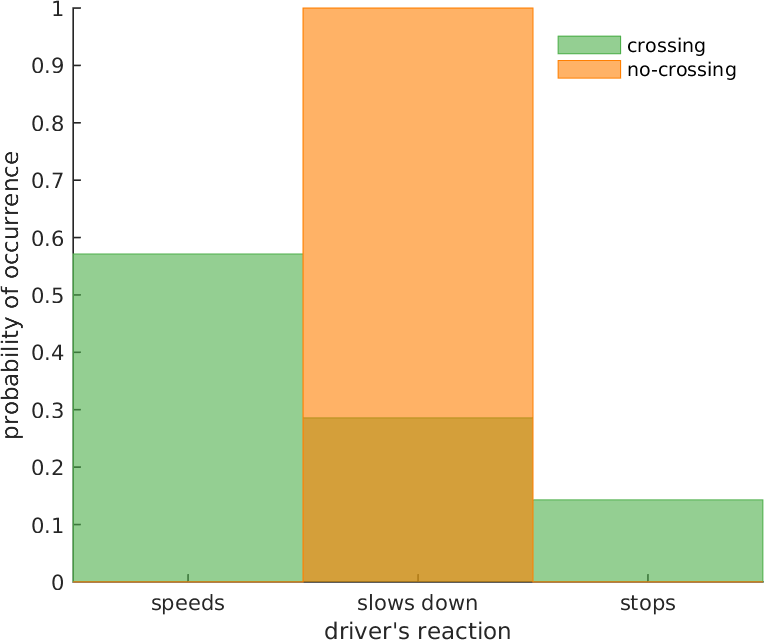}
\label{fig:crossing:driver_reaction_signal}}\\
\caption{Pedestrians crossing behavior at crosswalks with different properties.}
\vspace*{-\baselineskip}
\label{fig:crossing}
\end{figure}

Fig. \ref{fig:crossing:crossing_des} shows that pedestrians are less likely cross the street after communicating their intention if the crosswalk is not designated and more likely to cross if some form of signal or dedicated pathway is present.

To understand under what circumstances the crossing takes place in different crosswalks, we look at the driver's reaction to the pedestrian's intention of crossing. The driver's behavior can be grouped into \textit{speeds} (when the driver either maintains the current speed or speeds up), \textit{slows down} and \textit{stops}. 

Figs. \ref{fig:crossing:driver_reaction_no_des} and \ref{fig:crossing:driver_reaction_des} show that when there is no traffic signal present, in the majority of the cases pedestrians cross if the driver acknowledges their intention of crossing by slowing down or stopping. In few scenarios, the pedestrian still crosses the street even though the vehicle accelerates. In these cases either TTC is very high (average of 25.7 s) or the car is in a traffic congestion and the pedestrian anticipates that the car would shortly stop. Moreover, crossing also might not take place when the driver slows down or stops (even in the presence of a traffic signal) (see Fig. \ref{fig:crossing:driver_reaction_no_des} and\ref{fig:crossing:driver_reaction_signal}). In these cases  either the pedestrian hesitates to cross or explicitly (often by some form of hand gesture) yields to the driver.      

\section{CONCLUSIONS}
Pedestrians often engage in various forms of non-verbal communication with other road users. These include gazing, hand gesture, nodding or changing their behavior. At the point of crossing, in more than $90$\% of the cases pedestrians use some form of attention to communicate their intention of crossing. The most prominent form of attention (or primary communication) is looking in the direction of the approaching vehicles. The duration of looking also may vary depending on age of the pedestrian or time to collision. 

Other forms of explicit communication such as nodding or hand gesture were observed in $15$\% of the cases as a response to the driver's action and often were used to show gratitude, acknowledgement or to yield to the driver.

The crossing event does not always follow the first communication of intention. Crossing depends on additional factors such as the structure of the street (e.g. designated/non-designated, the width of the street), the driver's reaction to the communication or time to collision (how soon the driver arrives at the crosswalk).

Future work will include analysis of pedestrians' gait patterns with and without attention during the crossing. In addition, to better assess the nature of communication it would be beneficial to record driver's data such as driver's gestures, eye movements and any reaction that involves changing the state of the vehicle.\\

\bibliographystyle{IEEEtran}
\bibliography{references}

\end{document}